\documentclass[10pt,twocolumn,letterpaper]{article}

\usepackage[breaklinks=true,bookmarks=false]{hyperref}

\usepackage{iccv}
\usepackage{times}
\usepackage{epsfig}
\usepackage{graphicx}
\usepackage{amsmath}
\usepackage{amssymb}
\usepackage{times}
\usepackage{epsfig}
\usepackage{graphicx}
\usepackage{amsmath}
\usepackage{amssymb}
\usepackage{footnote}
\usepackage{dsfont}
\usepackage{comment}
\usepackage{blindtext}
\usepackage{morewrites}
\usepackage{makecell}
\usepackage{pifont}
\newcommand{\xmark}{\ding{55}}%
\newcommand{\cmark}{\ding{51}}%
\usepackage{multirow}

\usepackage[accsupp]{axessibility}  

\usepackage{footnote}
\usepackage[symbol]{footmisc}

\iccvfinalcopy 

\def\iccvPaperID{****} 

\ificcvfinal\fi

\begin{document}
\title{The Right to Talk: \\
An Audio-Visual Transformer Approach}

\author{Thanh-Dat Truong$^{1,*}$, Chi Nhan Duong$^{2, *}$, The De Vu$^{3}$, Hoang Anh Pham$^{3}$ \\ Bhiksha Raj$^{4}$, Ngan Le$^{1}$, Khoa Luu$^{1}$\\
$^{1}$CVIU Lab, University of Arkansas \quad $^{2}$Concordia University \\
$^{3}$VinAI Research \quad
$^{4}$Carnegie Mellon University\\
{\tt\footnotesize \{tt032, thile, khoaluu\}@uark.edu, dcnhan@ieee.org, \{v.devt, v.anhph18\}@vinai.io, bhiksha@cs.cmu.edu}
}
\maketitle
\ificcvfinal\thispagestyle{empty}\fi

\begin{abstract}

Turn-taking has played an essential role in structuring the regulation of a conversation. 
The task of identifying the main speaker (who is properly taking his/her turn of speaking) and the interrupters 
(who are interrupting or reacting to the main speaker's utterances) 
remains a challenging task. 
Although some prior methods have partially addressed this task, there still remain some limitations.  Firstly, a direct association of Audio and Visual features may limit the correlations
to be extracted 
due to different modalities.
Secondly, 
the relationship across temporal segments helping to maintain the consistency of localization, separation and conversation contexts is not effectively exploited. Finally, the interactions between speakers that usually contain the tracking and anticipatory decisions about transition to a new speaker is usually ignored.
Therefore, this work introduces a new Audio-Visual Transformer approach to the problem of localization and highlighting the main speaker in both audio and visual channels of a multi-speaker conversation video in the wild. The proposed method exploits different types of correlations presented in both visual and audio signals. The temporal audio-visual relationships across spatial-temporal space are anticipated and optimized via the self-attention mechanism in a Transformer structure.
Moreover, a newly collected dataset is introduced for the main speaker detection.
To the best of our knowledge, it is one of the first studies that is able to automatically localize and highlight the main speaker in both visual and audio channels in multi-speaker conversation videos.
\end{abstract}

\footnotetext{$^{*}$ denotes equal contributions}


\begin{table*} [t]
    \small
	\centering
	\caption{Comparisons of our proposed approach and other modeling methods. Sound Source Localization (SSL)}
	\footnotesize
\begin{tabular}{ >{\arraybackslash} m{3.2cm} c c c c c}
		\Xhline{2\arrayrulewidth}
		& \textbf{Ours} & LWTNet\cite{Afouras20b} & SyncNet \cite{chung2016out}  & SoundOfPixel \cite{Zhao_2018_ECCV}  & CocktailParty \cite{ephrat2018looking}  \\
		
		\Xhline{2\arrayrulewidth}
		
		
		\textbf{Goal} & \begin{tabular}{@{}c@{}} \textbf{Main Speaker} \\ \textbf{Highlight} \end{tabular} & \begin{tabular}{@{}c@{}} Active Speaker \\Highlight\end{tabular}   & SSL & Audio Separation & Audio Separation \\ 
		\hline
		\textbf{Temporal Model} & \textbf{Across-Segments} & Within-Segment  & Within-Segment & Within-Segment & Within-Segment \\ 
		\hline
		\textbf{People-Independent} & \cmark & \cmark & \xmark & \xmark & \cmark \\
		\textbf{Visual Context modeling (Visual-visual attention)} & \cmark & \xmark & \xmark & \xmark & \xmark \\

		\textbf{Audio-Visual Correlation} & \begin{tabular}{@{}c@{}} \textbf{Audio-Visual} \\ \textbf{Transformer} \end{tabular} & \begin{tabular}{@{}c@{}}Cosine \\ Distance\end{tabular}  & \begin{tabular}{@{}c@{}} Audio-Visual \\Synchronization \end{tabular} & \begin{tabular}{@{}c@{}} Feature \\Concatenation \end{tabular}& \begin{tabular}{@{}c@{}} Feature \\Concatenation \end{tabular} \\
		\Xhline{2\arrayrulewidth}
	\end{tabular}\label{tab:TenMethodSumm}
	\vspace{-4mm}
\end{table*}

\vspace{-4mm}
\section{Introduction}
\begin{figure}[t]
	\centering \includegraphics[width=1\columnwidth]{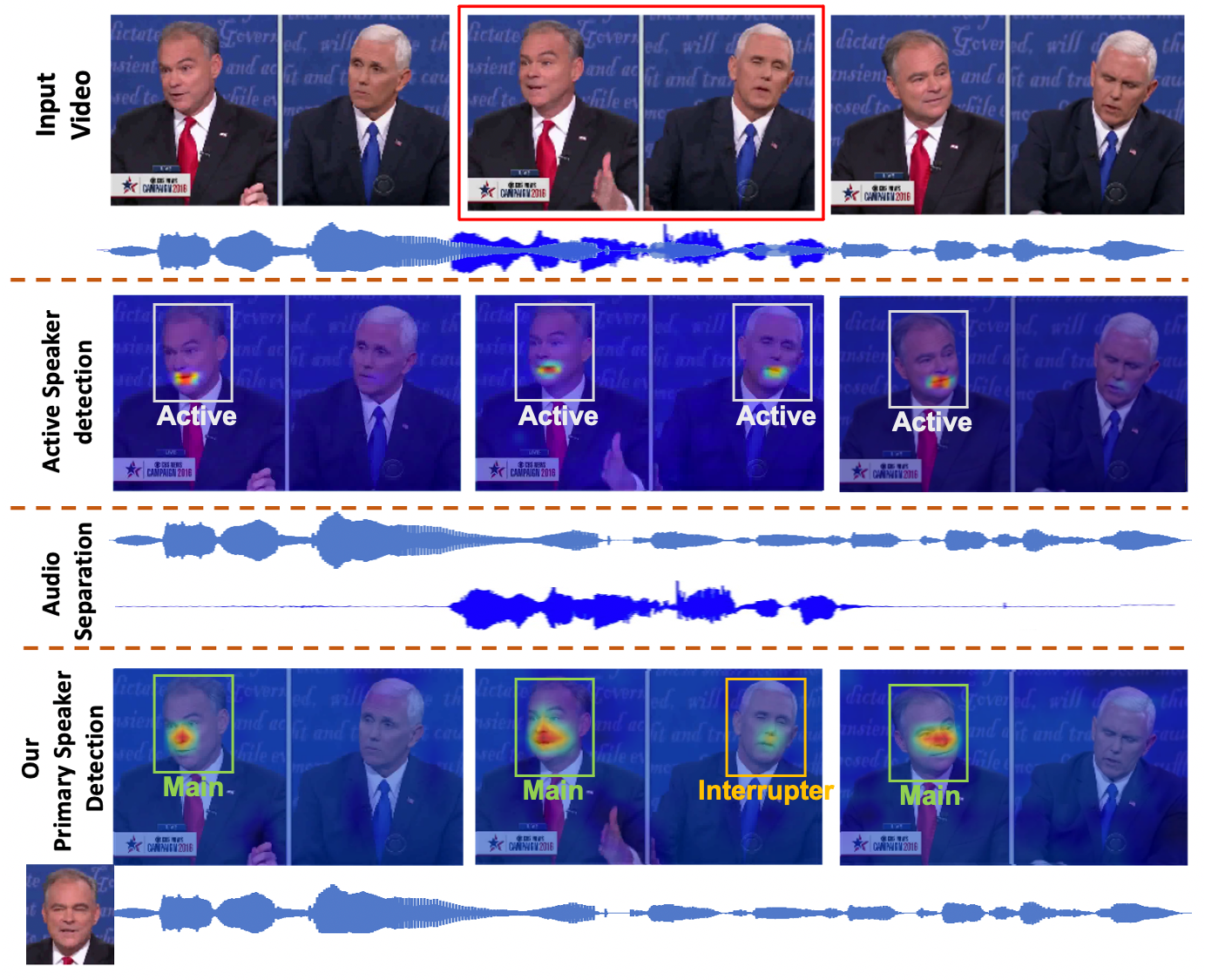}
	\small
	\caption{Given a multi-speaker video, our Audio-Visual Transformer can localize and highlight the main speaker in both visual and audio channels. (\textbf{Best viewed in color})}
	\label{fig:PrimarySpeakerDetection}
	\vspace{-6mm}
\end{figure}
Although human beings possess capabilities of localizing and separating sounds from noisy environments, we still have trouble following a conversation with noises, background voices, or interruptions from other speakers. 
Either with blind audio separation \cite{jin2009supervised,makino2007blind,reddy2007soft, radfar2007single,wang2005video} or visual-aid audio separation
\cite{afouras2019my, arandjelovic2017look, chung2019said, gabbay2018seeing, gao20192, harwath2018jointly, hu2019deep, liu2013source,khan2013speaker, owens2018learning, ramaswamy2020see, senocak2018learning,tian2018audio} approaches, this outlier separation task still remains a challenge in the wide conditions beyond the lab settings. The problem becomes especially harder when dealing with unknown numbers of speakers in an audio. Nachmani et al. \cite{nachmani2020voice} make a comparison between methods and show how hard it is to separate voices when the number of sound sources increases. 
Existing methods achieve high performance 
with inputs from multiple microphones. 
Some methods assume a clean set of single source audio examples are available for supervision \cite{afouras2018conversation, ephrat2018looking, zhao2019sound,Zhao_2018_ECCV}.
%
%
In practice, rather than solely trying to separate voices of all speakers in a conversation and determining ``\textit{who-spoke-when}'',
we tend to give more attentions to the \textbf{\textit{main speaker}}, i.e. \textit{who is on his/her turn of speaking and his/her talk is the main channel of communication}, and ignore the voices of remaining speakers, i.e. \textbf{\textit{interrupters}} or \textbf{\textit{listener}}, or background noises. 
Thus, 
an approach that highlights the main speaker in visual and audio channels would give new opportunities to popular applications such as auto-muting in a tele-conference or main speaker refocusable video generation.

Given a video of 
multi-speaker conversation, our goal is to learn an audio-visual model 
that enables the capabilities of both (1) localizing the main speaker; (2) true cancellation of audio sources of interrupters or background noises; and (3) automatically switching to a new subject when the speakers change their roles. The interruptions from other subjects and the background are considered as noises and removed.
In the scope of this work, we focus on \textbf{\textit{turn-taking conversation}} as the turn-taking mechanism has been commonly adopted for structuring conversation in social interactions. A subject is considered as the main speaker when he/she properly takes the turn of solo speaking and will continue the talk even after a simultaneous speech occurs \cite{cutrone2019profiling}.

\begin{figure*}[t]
	\centering \includegraphics[width=1.8\columnwidth]{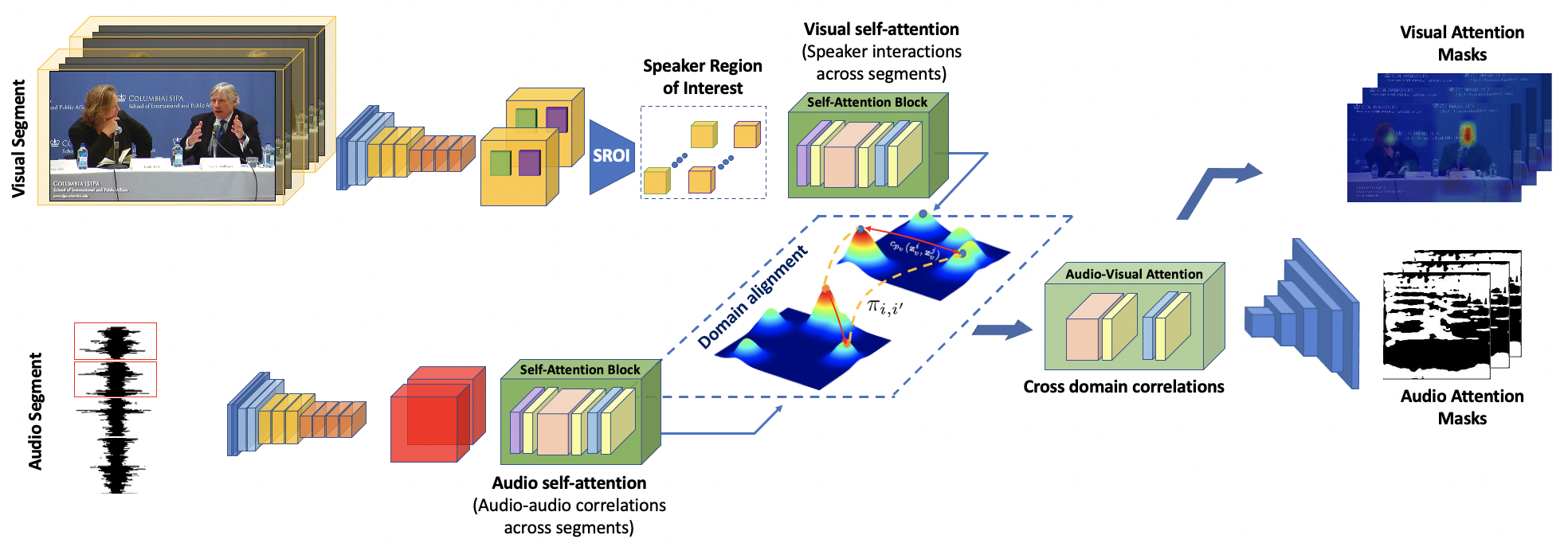}
	\small
	\caption{\textbf{Our Proposed Audio-Visual Transformer Framework.} Given a video segment set, the 
	context in the conversation is captured with
	three types of correlations, i.e. visual-visual, audio-audio, and audio-visual attentions. Then, the audio-visual attentional features are adopted for the main speaker localization and audio separation.
	(\textbf{Best viewed in color with 2x zoom in})
	}
	\label{fig:ProposedFramework}
	\vspace{-5mm}
\end{figure*}


Previous approaches have partially addressed 
this problem and can be divided into two categories, i.e.  \textit{audio-visual synchronization} \cite{arandjelovic2018objects, chung2016out,cutler2000look,owens2018audio,khosravan2019attention} and \textit{mix-and-separate} \cite{afouras2018conversation, 9054376, ephrat2018looking,gao2018learning,gao2019co, NEURIPS2020_7288251b, korbar2018co, zhao2019sound, Zhao_2018_ECCV, 10.1007/978-3-030-58610-2_4}. 
The former exploits the synchronization between audio and video frames within a specific time window to localize the image regions that are more sensitive to audio changes. Meanwhile, the latter 
learns to separate the speakers’ voices from a mix utterance based on audio and visual features. 
%
In both cases, there remain some limitations. Firstly, 
audio-visual relationships 
are extracted via a concatenation operator or the cosine distance metric. 
However, as audio and visual features distribute in two different latent spaces by their nature, 
these methods may not maximize correlations between the two feature domains.
Secondly, audio-visual relationships 
are only considered within a video segment, i.e. a short time window, while ignoring the ones across temporal segments, which helps to maintain the consistency of localization and separation, and \textit{contextual} main-subject switching.
Finally, the interactions between subjects in the temporal dimension to deliver accurate tracking and anticipatory decisions about transition to a new target are still ignored.

\noindent
\textbf{Contributions.} This work introduces a novel Audio-Visual Transformer approach, a cross-modality temporal-based computer vision algorithm, to highlight main speaker in both audio and visual channels 
(Fig. \ref{fig:PrimarySpeakerDetection}).
The contributions of this work are four-fold. 
(1) The proposed approach exploits various correlations presented in visual and audio signals 
including ``virtual'' interactions between speakers in a video scene and relationships between visual and auditory modalities.
(2) Rather than extracting audio-visual correlations within a video segment, relationships across segments are further exploited via a temporal self-attention mechanism in the proposed Transformer structure. 
This helps to engage the contextual information and enhance attentions with longer context so that the main speaker can be robustly identified.
(3) A Cycle Synchronization Loss is introduced to learn the main speaker localization in a self-supervised manner.
(4) A newly dataset\footnote{The dataset and implementation will be publicly available at \url{https://github.com/uark-cviu/Right2Talk}} is collected for the main speaker detection.
To the best of our knowledge, it is one of the first works that is able to automatically localize and highlight the main speaker in multi-speaker conversation videos on both visual and audio channels (Table \ref{tab:TenMethodSumm}).



    


 


\section{Related Work}

\noindent
\textbf{Active Speaker Localization.}  
This problem aims to localize the sources of sounds in a given  
video. 
Some early methods \cite{barzelay2007harmony,hershey1999audio, izadinia2012multimodal,kidron2005pixels, senocak2018learning} localize the sources of human voice in a video using statistical models and audio-visual correlations. 
Fisher et al. \cite{fisher2000learning} introduces 
a multi-media fusion method in a complex domain to capture latent audio-visual relationships.
%
%
Later, deep learning approaches \cite{chung2016out,owens2018audio} come into place and exploit the synchronization between visual and audio signals to find the regions in the images that are sensitive to the audio features.
%
%
Afouras et al. \cite{Afouras20b} propose LWTNet that extends the synchronize cues with optical flow technique to extract and track audio-visual objects for the localization process. 
%
%
Unlike previous 
approaches, our work goes one step further by taking into account the \textit{context} of audio-visual features that presents across video segments. 
Our method does not naively localize all the areas containing voice, but it is able to notice the sound and highlight the location of the \textit{main speaker} in a conversation.

\noindent
\textbf{Speaker Audio Separation.}
Prior methods \cite{jin2009supervised, makino2007blind,reddy2007soft} use audio-only features, i.e characteristics of voice, to resolve this problem. Hershey et al. \cite{hershey2016deep} formulate this task as a clustering problem where the objective is to learn an embedding for each time-frequency element in the spectrogram, such that each embedding cluster associates with a voice of a subject.
%
Zhao et al. \cite{zhao2019sound, Zhao_2018_ECCV} detect objects in one or multiple frames and use their appearance and motion to differentiate sounds of objects. 
%
Gao et al. \cite{gao2019co} propose a co-separation training objective to learn audio-source separation from unlabeled videos containing multiple sources of sounds. 
%
Ephrat et al. \cite{ephrat2018looking} contribute a large-scale dataset, namely AVspeech, and propose an end-to-end audio-visual architecture. 
Afouras et al. \cite{afouras2018conversation} propose to use the lip regions and consider both audio magnitudes and phases. 
%
The aforementioned methods ignore the \textit{context} of the video which is a very important cue for the network to improve the quality of separated voices. 
Our novel architecture is proposed to 
to gain the \textit{context} information . 

\section{The Proposed Method}
This work focuses on turn-taking conversations composing talking turns. 
The length of each turn is flexible according to the conversation's context and contents.
%
Let $\mathbf{x} \in \mathcal{X}$ be a multi-speaker conversation video consisting of a \textit{visual component} $\mathcal{V}$ (a sequence of RGB frames) and an \textit{audio component} $ \mathcal{A}$ (a mixed audio of one or multiple speakers).


\subsection{The Turn-taking Conversation} \label{sec:InTurnConversation}
With turn-taking regulations, a conversation $\mathbf{x}$ can be decomposed into speaking turns, i.e. $\mathbf{Turn}^h, h=1,...,H$ where $H$ is the number of turns in $\mathbf{x}$. Although many speakers can have their voices overlapped during a speaking turn in either cooperative or competitive manner, the role of each speaker can be classified into two groups.

\vspace{1mm}
\noindent
\textbf{\textit{Main Speaker} $S_m^h$. } A subject $S$ is 
the main speaker of $\mathbf{Turn}^h$ when he/she carries the conversation and drives it forward \cite{cutrone2019profiling}. Even an interruption (i.e. simultaneous speech) occurs, the subject will continue to solo speak until the end of the turn. Thus, a long solo speaking of $S$ during $\mathbf{Turn}^h$ can provide an  indication for the main speaker role. 

\begin{figure}[t]
	\centering \includegraphics[width=1\columnwidth]{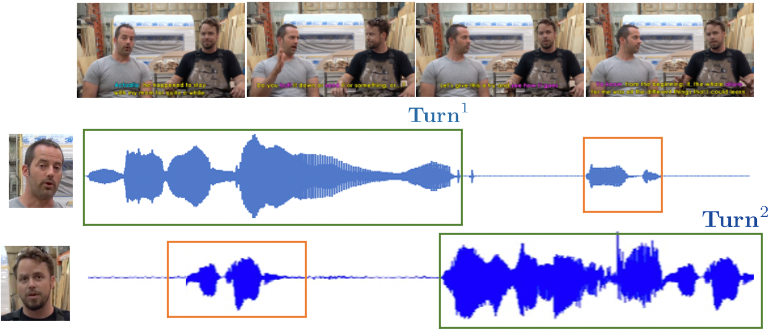}
	\small
	\caption{\textbf{Turn-taking Conversation.} The conversation is decomposed in to turns where each speaker acts as the main speaker (green box) of a turn, and the other speakers are considered as interrupter or listener (orange box) during that turn.}
	\label{fig:TurnTakingConversation}
	\vspace{-5mm}
\end{figure}

\vspace{1mm}
\noindent
\textbf{\textit{Interrupter or Listener} $S_I^h$. } 
An interrupter or Listener is the one who has reactions or comments to the main speaker's utterances. These reactions usually occur in a short time window during $\mathbf{Turn}^h$ and end up with the continuation of the main speaker's talk.
%
When the interrupter continues to solo speak after  a simultaneous speech, a turn changing of the main speaker occurs. Fig. \ref{fig:TurnTakingConversation} illustrates an example of speakers' roles in a turn-taking conversation.



\vspace{-0.5mm}
\subsection{Problem Definition}
\vspace{-0.5mm}
Rather than decomposing $\mathbf{x}$ into $\{\mathbf{Turn}^h\}_1^H$, we propose to present  $\mathbf{x}$ as a composition of $K$ segments $\mathbf{Seg}^k = \{\mathbf{v}^k,\mathbf{a}^k \}, k=1,...,K$, $\mathbf{v}^k \in \mathcal{V}$ and $\mathbf{a}^k \in \mathcal{A}$. The $h$-th speaking turn $\mathbf{Turn}^h$ consists of one or multiple segments, i.e. $\mathbf{Turn}^h = \{\mathbf{Seg}^k\}_{h_{start}}^{h_{end}}$ where $h_{start}$ and $h_{end}$ mark the indices of the starting and ending time of $h$-th turn, respectively. 
Let $S_m^k$ be the main speaker and $S_I^k$ be the interrupters of $\mathbf{Seg}^k$ in the conversation. We have $S_m^k \equiv S_m^h$ and $S_I^k \equiv S_I^h$ when $\mathbf{Seg}^k \in \mathbf{Turn}^h$. Then, the goal is to extract the location and clean voice of $S_m^k$ for each $\mathbf{Seg}^k$ in the conversation.
Formally, the objectives are to learn the visual location map $\mathbf{M}_v^k$ and audio mask $\mathbf{M}_a^k$ of $S^k_m$ as.
%
%
\begin{equation} \label{eqn:object_learning_visual}
\footnotesize
\begin{split}
    \mathbf{M}_v^{k*} = \arg \min_{\mathbf{M}_v^{k}}
   \Big[ -&\log P\left(\mathbf{M}_v^{k}[Loc(S_m^k, \mathbf{v}^k)]|\mathbf{Seg}^{1:k}\right) \\
 + &\log P\left(\mathbf{M}_v^{k}[Loc(S_I^k, \mathbf{v}^k)]|\mathbf{Seg}^{1:k}\right) \Big]
\end{split}
\end{equation}
\begin{equation} \label{eqn:object_learning_audio}
\footnotesize
    \mathbf{M}_a^{k*} = \arg \min_{\mathbf{M}_a^{k}} \| \mathbf{M}_a^{k} \odot Spec(\mathbf{a}^k) - Spec(\mathbf{a}_{S_m^k})\|_1
\end{equation}
where $\textit{Loc}(S_m^k, \mathbf{v}^k)$ is the location of $S_m^k$ in $\mathbf{v}^k$;
$Spec(\cdot)$ is the spectrogram conversion operator; $\odot$ is the Hadamard product;
and $\mathbf{a}_{S_m^k}$ is the clean voice of  $S_m^k$. 
$\mathbf{Seg}^{1:k}$ denotes the temporal information provided from the beginning to the $k$-th segment of the video $\mathbf{x}$. The conditional term indicates the temporal constraint 
being considered.


To effectively estimate 
$\mathbf{M}_v^k$ and 
$\mathbf{M}_a^k$,
we propose an Audio-Visual Transformer approach (see Fig. \ref{fig:ProposedFramework}) consisting of three learning stages: (1) Learning the context with visual and audio self-attention; (2) Audio-Visual Correlation Learning; and (3) Main speaker localization and audio separation with Conversation Grammar. The proposed Audio-Visual Transformer is formulated via 
$\{D,\phi, E_v, E_a\}$ as:
\begin{equation}
\footnotesize
\begin{split}
    G &= [D \circ \phi] (\mathbf{z}_v ,\mathbf{z}_a)\\
    \mathbf{z}^k_v &= E_v(\mathbf{v}^k| \mathbf{Seg}^{1..k})\\
    \mathbf{z}^k_a &= E_a(\mathbf{a}^k| \mathbf{Seg}^{1..k})\\
\end{split}
\end{equation}
where $E_v$ and $E_a$ map 
$\mathbf{v}^k$ and $\mathbf{a}^k$ to their
latent representations; and $\circ$ is the functional composition.
$\phi$ is the projection function to the shared representation space where these modalities are comparable. 
$D$ maps these deep representations to the audio-visual mask of the main speaker.

\vspace{-0.5mm}
\subsection{Visual and Audio Contextual Learning}
\vspace{-0.5mm}
Besides prior works on temporal learning \cite{7780991, 10.1007/s11263-018-1113-3, 8953561, 10.1007/s11263-019-01165-5, 8237665, Duong_2020_CVPR, fi13070179}, in this work, given a conversation, contextual information can be extracted from visual and audio signals, i.e. \textit{visual-visual} and \textit{audio-audio} correlations \textit{across video segments}. While the former assists to track behaviors and interactions of each speaker over the spatial-temporal dimension, the latter provides more cues about the conversation flow,  i.e. when and how the main speaker switch his/her role. 
For example, the higher audio-audio correlation between two or more segments is, the lower the possibility is of the main speaker being switched. 
Thus, these cross-segment correlations can implicitly embed turn changing signals of the main speaker, and help to avoid the stage of pre-decomposing $\mathbf{x}$ into speaking turns $\mathbf{Turn}^h$. Even when interrupters dominate the main speaker's voice in a certain segment, this ``temporal-based'' correlation can exploit the relations to previous segments and identify the main speaker. 
We 
model the contextual correlations via two encoder structures with a self-attention mechanism before embedding their cross-domain correlations.

\vspace{-0.5mm}
\subsubsection{Visual-Visual Self-Attention.} \label{sec:Visual_Self_Att}
Given a sequence of segments $\{\mathbf{v}^k\}_1^K$, the visual encoder $E_v$ consists of three main functions, i.e. feature embedding, self-attention, and feature refinement with attention. Particularly, each $\mathbf{v}^k$ is firstly embedded into a deep feature embedding via $F_{v}: \mathcal{V} \mapsto \mathcal{F}_v$ as $\mathbf{f}^k_v = F_v(\mathbf{v}^k)$.

\vspace{1mm}
\noindent
\textit{\textbf{Speaker Region of Interest (SROI).}}
Rather than embedding each $\mathbf{v}^k$ into a single feature for attention computation, we propose to project $\mathbf{f}^k_v$ into regions of interest where each region represents a speaker's location 
in a visual segment and learn the correlations among them. Particularly, let $\mathbf{b}=\{ \mathbf{b}_i^k\}, i=1..N, k=1..K$ where $\mathbf{b}_i^k \in \mathcal{B}  \subset \mathbb{R}^4$ denotes the location of $i$-th speaker in $\mathbf{Seg}^k$, and $N$ is the number of speakers. The projection function $R: \mathcal{F}_v \times \mathcal{B} \mapsto \mathcal{F}_{v}$ is defined as $\mathbf{f}^{k,i}_{v} = R(\mathbf{b}^k_i, \mathbf{f}_v^k)$. We adopt ROI Align \cite{He_2017_ICCV} for the function $R$.
There are two approaches to obtain $\mathbf{b}$ and $N$, i.e. face detection and block decomposition. The former adopts a face detection 
to extract faces 
in all segments. 
The latter uniformly decomposes a visual segment into $N = n \times n$ blocks for $\mathbf{b}$. While face detection approach tends to give more direct focus on
face regions, our experiments show that block decomposition can provide attentions to regions of face track of the same speaker across segments.


\begin{figure}[t]
    \centering
    \includegraphics[width=0.46\textwidth]{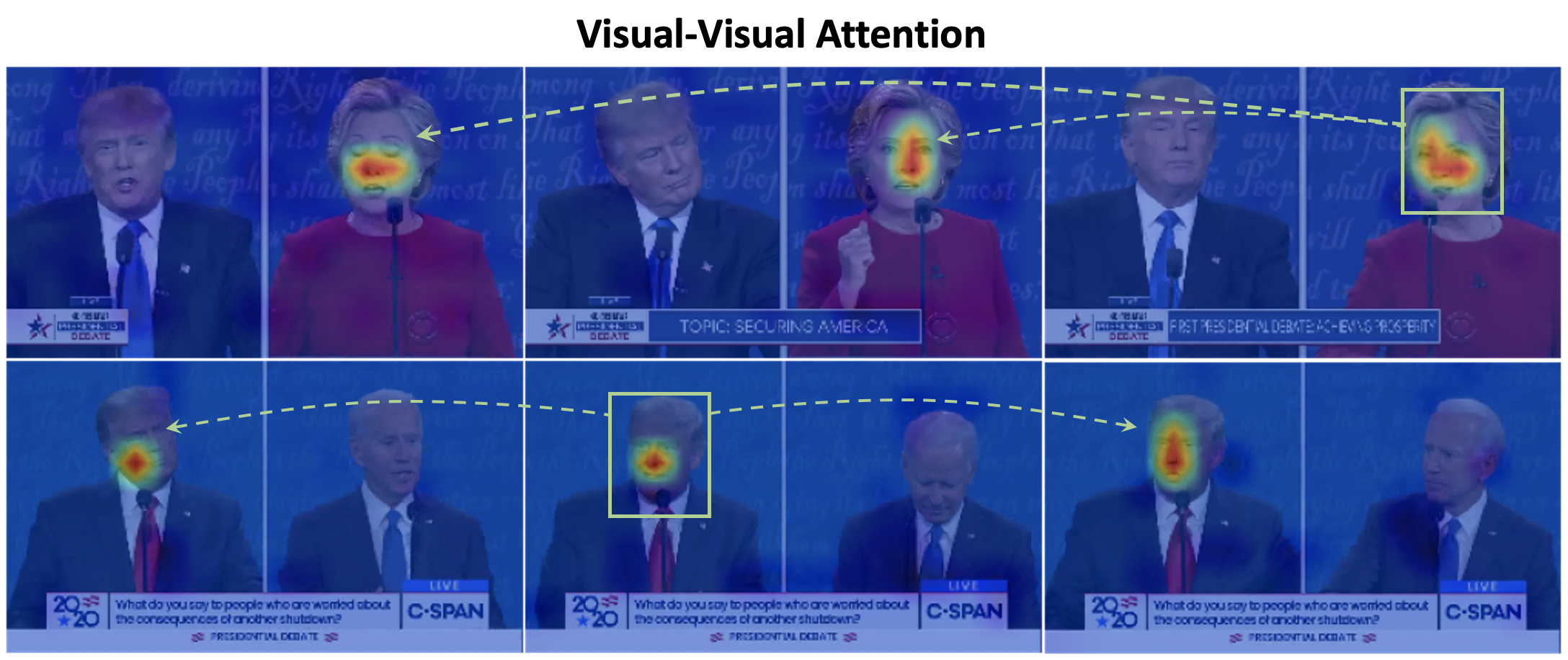}
    \caption{\textbf{Visual-Visual Attention.} The attention masks across video segments corresponding to the speaker in the green box. This type of attention can help to track the behaviors and interactions of each speaker over the  spatial-temporal  dimension.} 
    \label{fig:Visual_Visual_attention}
    \vspace{-4mm}
\end{figure}

\vspace{1mm}
\noindent
\textit{\textbf{Virtual Interaction Attention.}}
Given a feature set $\mathbf{f}^{k,i}_{v}$, the visual-visual context across the spatial-temporal dimension can be expressed as building a dynamic dictionary per feature set with three basic attention based elements
\cite{Nguyen_2021_CVPR, dyglip_cvpr21}, i.e. \textit{key, query, value}. While \textit{key} and \textit{query} are trained to support the dictionary look-up process where query feature is highly correlated to its matching key and dissimilar to others, \textit{value} represents a discriminative feature for each speaker. Particularly, the self-attention set $\{\mathbf{k}^{k,i}_v, \mathbf{q}^{k,i}_v, \mathbf{v}^{k,i}_v \}$ is extracted via three learnable projections $\{\Omega^Q_{v}, \Omega^K_{v}, \Omega^V_{v}\}$ as ${\mathbf{q}^{k,i}_{v}} = \Omega^Q_{v}(\mathbf{f}^{k,i}_{v}); {\mathbf{k}^{k,i}_{v}} = \Omega^K_{v}(\mathbf{f}^{k,i}_{v}); {\mathbf{v}^{k,i}_{v}} = \Omega^V_{v}(\mathbf{f}^{k,i}_{v})$. 
The visual correlation among speakers can be defined as.
\begin{equation} \label{eqn:att_vis_vis}
\footnotesize
\alpha_{v}^{ki,k'j} = \sigma\left(\mathbf{q}^{k, i}_v ({\mathbf{k}^{k', j}_v})^{\top}/ \sqrt{d}\right)
\end{equation}
where $d$ is the feature dimension, $k'$ is a segment indexing variable. 
We consider the attention as a probability distribution that illustrates the responsive attention among speakers. Therefore, the softmax function can be adopted for $\sigma (\cdot)$.

\vspace{1mm}
\noindent
\textit{\textbf{Feature Refinement with attention.}}
With these correlations, the visual self-attention
among speakers 
allows every speaker correlates to all other speakers through the spatial-time dimension. Then, the virtual interaction over speakers is explicitly embedded to their representations as.

\begin{equation}
\footnotesize
    \mathbf{z}^{k, i}_v = \eta_{v}\left(\mathbf{f}^{k, i}_v + \sum_{k'=1}^K\sum_{j=1}^N\alpha^{ki, k'j}_v\mathbf{v}^{k,i}_v\right)
\end{equation}
where $\eta_v$ is the a residual-style MLP.
Throughout this process, the features of each speaker in one visual segment can interactively embed in their latent representation the correlations with those of the same speaker in other segments as well as other speakers of the same segment. Fig. \ref{fig:Visual_Visual_attention} illustrates the attention mask across video segments corresponding to the speaker in the green box.

\begin{figure}[!t]
    \centering
    \includegraphics[width=0.45\textwidth]{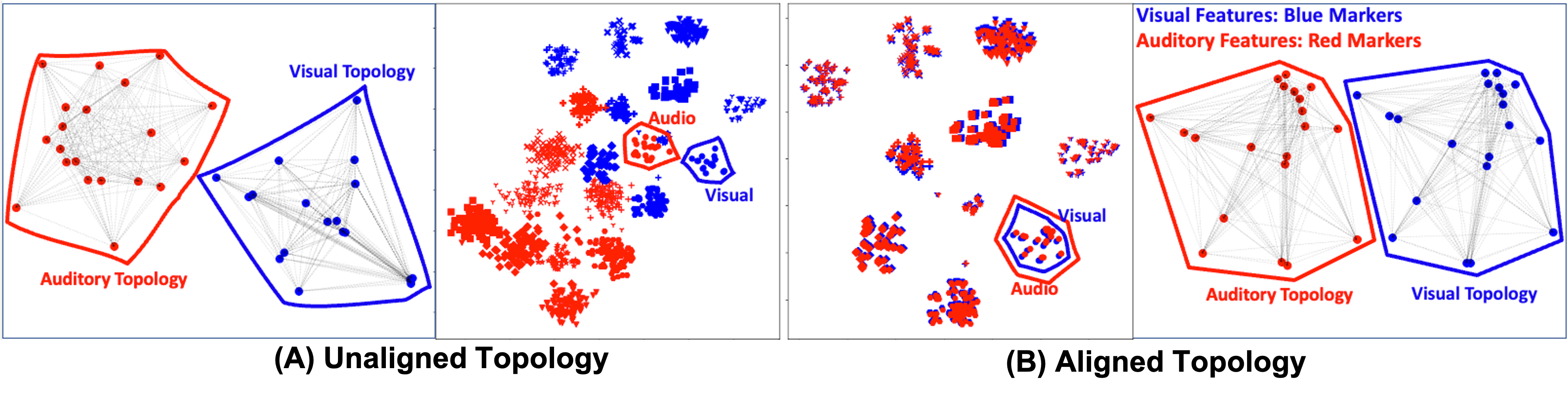}
    \caption{Topology of audio and visual feature domains.}
    \label{fig:topo}
    \vspace{-4mm}
\end{figure}

\vspace{-2mm}
\subsubsection{Audio-Audio Self-Attention}
Similar to the virtual interaction, the audio self-attention is modeled as the correlation among audio segments. 
Particularly, let $F_a: \mathcal{A} \to \mathcal{F}_a$ be an audio embedding function that extracts audio latent representation for audio segments $\{\mathbf{a}^k\}_1^K$ and $\mathbf{f}^k_a = F_a(\mathbf{a}^k)$.
The audio self-attention correlation among segments can be computed as in Eqn. \eqref{eqn:att_aud_aud}.
\begin{equation} \label{eqn:att_aud_aud}
\footnotesize
    \begin{split}
        {\mathbf{q}^{k}_{a}} = \Omega^Q_{a}(\mathbf{f}^{k}_{a}); 
        {\mathbf{k}^{k}_{a}} &= \Omega^K_{a}(\mathbf{f}^{k}_{a});  
        {\mathbf{v}^{k}_{a}} = \Omega^V_{a}(\mathbf{f}^{k}_{a}) \\
        \alpha_{a}^{k,k'} &= \sigma\left(\mathbf{q}^{k}_a({\mathbf{k}^{k'}_a})^{\top}/\sqrt{d}\right)\\
    \mathbf{z}^k_a &= \eta_a(\mathbf{f}^k_a + \sum_{k'=1}^K \alpha_a^{k, k'}\mathbf{v}^{k'}_a)
    \end{split}
\end{equation}
%
The extracted audio feature of each segment is able to embed the correlation with other audio segments through time.

\subsection{Audio-Visual Correlation Learning}
The audio-visual correlations 
are computed from features of two different domains. As \textit{\textbf{topologies}}, i.e. \textit{how 
features distributed in the latent space and the correlations among its features} (see Fig. \ref{fig:topo}), of these modalities may differ significantly, directly associating these features for correlation learning is not efficient.
One solution is to set up two encoders $E_v$ and $E_a$ extracting features from latent spaces of the same dimension to leverage the domain differences. However, the topology difference between these modalities may still present. To mitigate this issue, we 
align two domains' topologies before learning the correlations between their features. By this way, $\mathbf{z}^k_a$ and $\mathbf{z}^k_v$ are well aligned and their correlations can be fully exploited.


\begin{figure}[!t]
    \centering
    \includegraphics[width=0.45\textwidth]{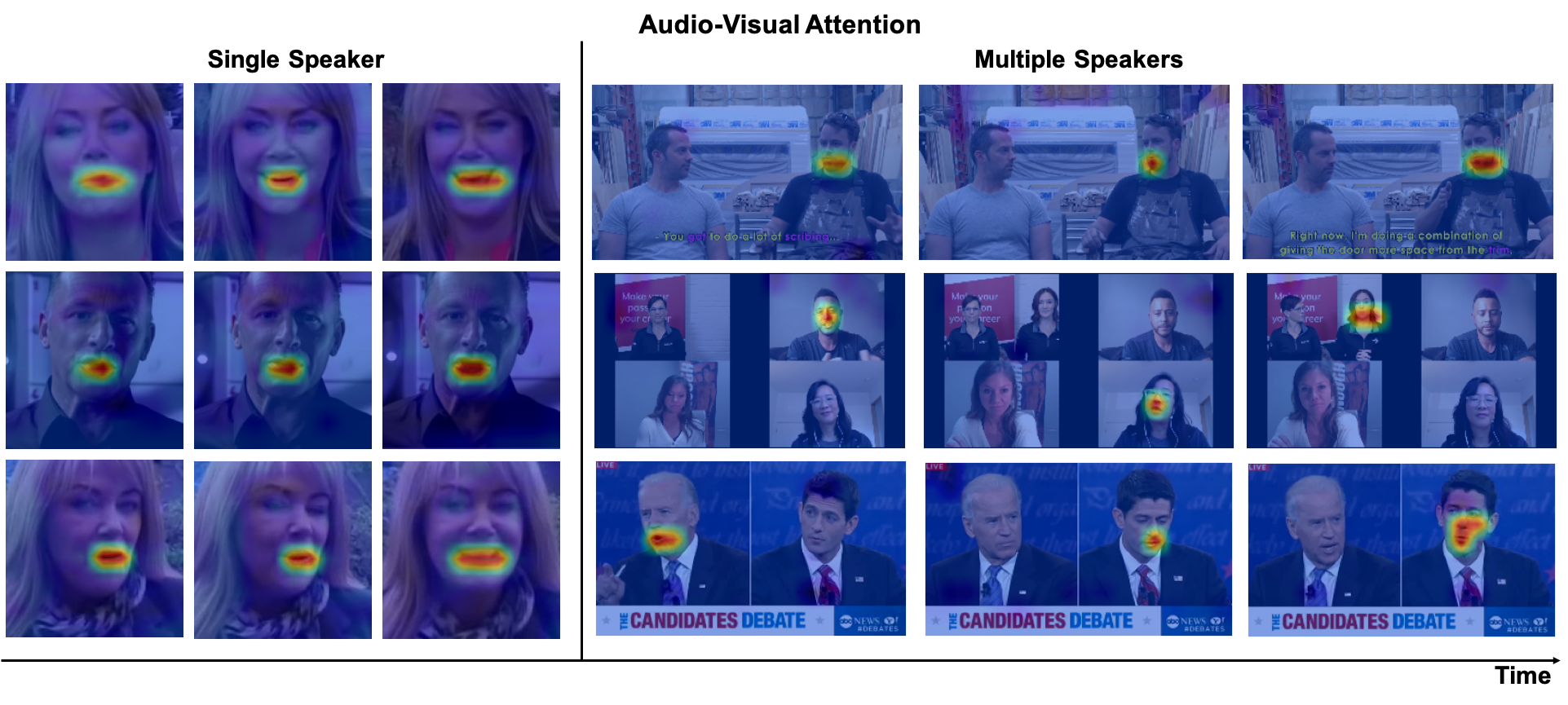}
    \caption{\textbf{Audio-Visual Attention To The Main Speaker.} The audio-visual attention mask across segments illustrates the response of audio to the visual.
    (\textbf{Best viewed in color})}
    \label{fig:attnetion_visual_audio_speaker}
    \vspace{-5mm}
\end{figure}

\vspace{1mm}
\noindent
\textit{\textbf{Cross Domain Alignment as Optimal Transport (OT) Problem.}}
We present the distributions of visual and audio features by two distributions $p_v$ and $p_a$ where $\mathbf{z}_v \sim p_v(\mathbf{z}_v)$, and $\mathbf{z}_a \sim p_a(\mathbf{a}_a)$; and propose a two-stage alignment process: (1) Sample association between visual and audio samples via transport function $\pi$ and (2) Topology synchronization.
Formally, let $\pi$ be the transport function where $\pi_{i,i'}=\pi(\mathbf{z}_v^i, \mathbf{z}_a^{i'})$ indicates the probability of association between a visual sample $\mathbf{z}_v^i$ and an audio sample $\mathbf{z}_a^{i'}$. In addition, let $c_{p_v}(\cdot, \cdot)$ and $c_{p_a}(\cdot, \cdot)$ be the cost functions defined as the distance between two samples in visual and audio spaces, respectively. 
The alignment process is formulated with Gromov-Wasserstein distance as shown in Eqn. \eqref{eqn:GW_distance}.
\begin{equation} \label{eqn:GW_distance} 
\footnotesize
\begin{split}
    \mathcal{L}_{align} = GW^2(c_{p_v}, c_{p_a}, p_v, p_a) = \min_{\pi \in \Pi(p_v, p_a)}J(c_{p_v}, c_{p_a}, \pi)\\
    J(c_{p_v}, c_{p_a}, \pi) = \sum_{i,j,i',j'}|c_{p_v}(\mathbf{z}_v^i, \mathbf{z}_v^j) - c_{p_a}(\mathbf{z}_{a}^{i'}, \mathbf{z}_{a}^{j'})|^2\pi_{i,i'}\pi_{j,j'}
\end{split}
\raisetag{30pt}
\end{equation}
Intuitively, minimizing $J(c_{p_v}, c_{p_a}, \pi)$ aims at finding an appropriated association (i.e. via $\pi$) between samples in the two domains as well as minimizing the topology difference between them (i.e. via $c_{p_v}, c_{p_a}$). Notice that directly solving Eqn. \eqref{eqn:GW_distance} is costly due to the non-convex Quadratic Problem with the time complexity is $O(n^3)$. Therefore, we adopt the the sliced approach \cite{vay_sgw_2019} for a fast computation of $\mathcal{L}_{align}$. Fig. \ref{fig:topo} illustrates visual (blue points) and auditory features (red points) extracted from 500 clip segments of 10 different speakers (e.g. denoted by different markers) and projected into the 2D space using t-SNE method. 
Thanks to $\mathcal{L}_{align}$ in the alignment stage, visual and auditory features are brought into similar distributions (Fig. \ref{fig:topo} (B) (left)) with more aligned feature distributions (Fig. \ref{fig:topo} (B) (right)).

\vspace{1mm}
\noindent
\textit{\textbf{Audio-Visual Correlation.}}
With the aligned visual and audio features, we further adopt similar attention mechanism to learn the associations between the visual features of each SROI $\mathbf{z}^{k,i}_{v}$  and the audio features $\mathbf{z}^k_a$ in each segment as. 
\begin{equation} \label{eqn:visual_att_aud_aud}
\footnotesize
    \begin{split}
        {\mathbf{q}^{k}} = \Omega^Q(\mathbf{z}^{k}_{a}); 
        {\mathbf{k}^{k, i}} &= \Omega^K(\mathbf{z}^{k, i}_{v});  
        {\mathbf{v}^{k, i}} = \Omega^V(\mathbf{z}^{k, i}_{v}) \\
        \alpha^{k,k'i} &= \sigma\left(\mathbf{q}^{k}({\mathbf{k}^{k', i}})^{\top}/\sqrt{d}\right)\\
        \mathbf{z}^k = \phi(\mathbf{z}^k_a, \mathbf{z}^k_v) &= \eta\left(\mathbf{z}^k_a + \sum_{k'=1}^K\sum_{i=1}^N \alpha^{k, k'i}\mathbf{v}^{k', i}\right)
    \end{split}
\end{equation}

%
%
The attention matrix 
assesses how much an audio responds to an SROI in the spatial-temporal dimension. A high response indicates a high correlation between the audio and the speaker associated with that SROI. This association embeds the probability of a speaker to be an active speaker of the audio segment. Fig. \ref{fig:attnetion_visual_audio_speaker} illustrates audio-visual attentions in both single and multiple speaker conversation.
\subsection{Main Speaker Localization and Audio Separation with Conversation Grammar} \label{sec:LearningProcess}
Given the audio-visual attentional features $\mathbf{z}^k$ from previous step, the audio-visual masks are computed as follows,
\begin{equation}
\footnotesize
\begin{split}
    \mathbf{M}_v^k(x,y) &= \begin{cases}
    \alpha^{k,ki}       &  \text{if } (x,y) \in \mathbf{b}_i^k, i=1..N\\
    0  & \text{otherwise}
  \end{cases} \\
  \mathbf{M}_a^k &= D(\mathbf{z}^k)
\end{split}
\end{equation}
where $D$ is a learnable decoder that maps $\mathbf{z}^k$ to the target audio mask; and $\alpha^{k,ki}$ denotes the correlation score between the audio and SROI of $i$-th speaker in the $k$-th segment.
The objective functions of Eqns. \eqref{eqn:object_learning_visual} and  \eqref{eqn:object_learning_audio} to learn $\mathbf{M}_a^k$ and $\mathbf{M}_v$ can be reformulated as follows,
\begin{equation}
\footnotesize
\begin{split}
    \mathcal{L}_{visual} &= \mathbb{E} \left[-\log \frac{e^{\alpha^{k,ki}}}{e^{\alpha^{k,ki}} + \sum_{j\neq i}e^{\alpha^{k,kj}}}\right]\\
    \mathcal{L}_{audio}& = \mathbb{E} \left[|| \mathbf{M}^k_a \odot Spec(\mathbf{a}^k) - Spec(\mathbf{a}_{S^k_m})||_1\right]
\end{split}
\end{equation}
Intuitively, on one hand, $\mathcal{L}_{audio}$ optimizes the model toward voice of the target (Main) speaker. On the other hand, $\mathcal{L}_{visual}$ aims at increasing the correlations between the audio and SROI of the target speaker, while reducing the correlation with other SROIs in spatial-temporal dimensions.

\begin{figure}[t]
	\centering \includegraphics[width=0.8\columnwidth]{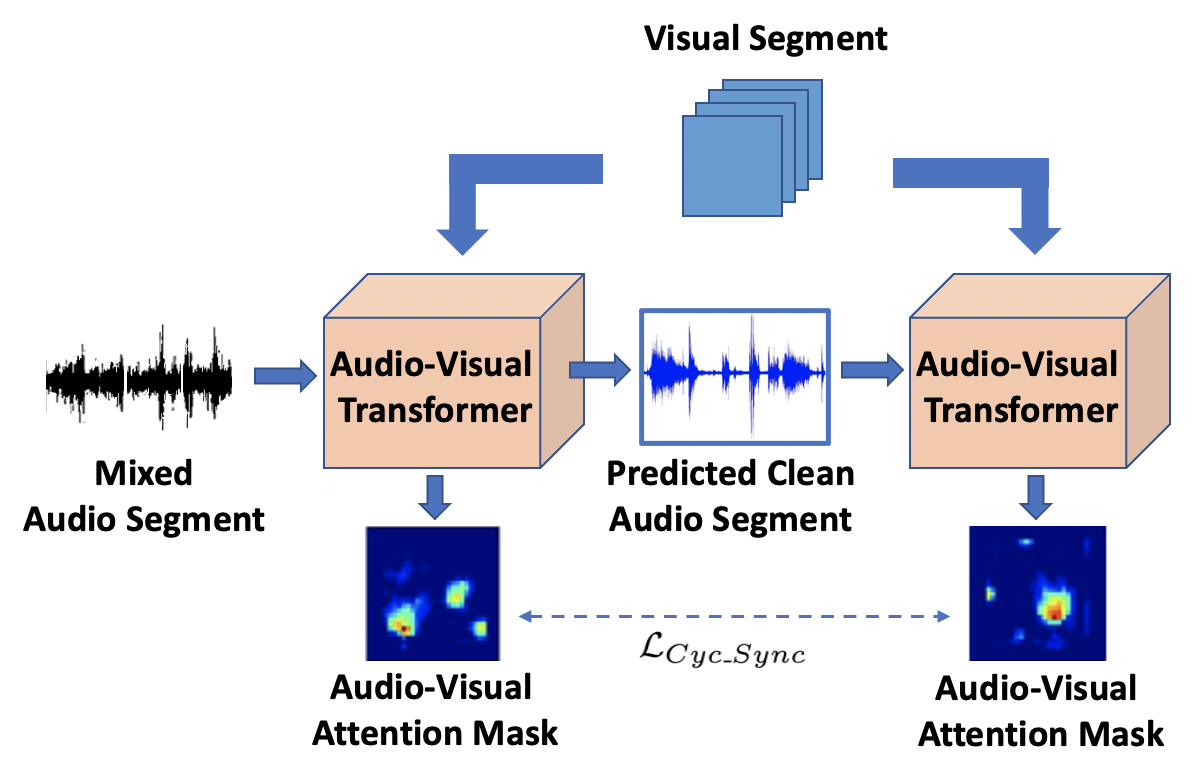}
	\small
	\caption{\textbf{Cycle Synchronization Loss.}}
	\label{fig:Loss_Cyc_Sync}
	\vspace{-5mm}
\end{figure}

\vspace{1mm}
\noindent
\textit{\textbf{Self-supervised Learning.}}
While our goal is to develop a self-supervised model that learns to localize the main speaker, the ground truth location for Main speaker is absent during the training stage. Therefore, we further propose a self-supervised version of $\mathcal{L}_{visual}$ , namely \textit{Cycle Synchronization Loss}, defined as follows,
\begin{equation}
\footnotesize
\begin{split}
    \mathcal{L}_{Cyc\_Sync} &= \mathbb{E} \left[ ||\alpha^{k,ki} - \hat{\alpha}^{k,ki}_{\text{max}}||_1\right]\\
    \hat{\alpha}^{k,ki}_{\text{max}} &= \begin{cases}
    \hat{\alpha}^{k,ki}       &  \text{if } (x,y) \in \mathbf{b}_{i^*}^k\\
    0  & \text{otherwise}
  \end{cases}; 
  i^* = \arg \max_i \hat{\alpha}^{k,ki}
\end{split}
\end{equation}
where $\hat{\alpha}^{k,ki}$ is the correlation between the predicted clean voice of the target speaker. The intuition of $\mathcal{L}_{Cyc\_Sync}$ is illustrated in Fig. \ref{fig:Loss_Cyc_Sync} where the goal is to penalize the consistency between two terms: (1) the correlations of the input (mix) voices $\mathbf{a}^k$ and the visual component; and (2) the maximum correlations of the predicted (clean) voice of the target and the visual component.
As clean voice is of a single speaker and it reflects similar linguistic content as visual features of the target speaker, its correlations with the visual component can efficiently act as the guidance for localization process.
Moreover, by considering only the maximum $\hat{\alpha}^{k,ki}$ in $\hat{\alpha}^{k,ki}_{\text{max}}$, the correlation between the visual of other speakers (i.e. interrupter) and audio is also minimized.

\vspace{1mm}
\noindent
\textit{\textbf{Learning with Conversation Grammar.}} We adopt the mix-and-separate strategy \cite{ephrat2018looking, Zhao_2018_ECCV} to obtain the ground truth for audio separation task of $\mathcal{L}_{audio}$ and further extend it with two types of Conversation Grammar, i.e. cooperative and competitive modes.  
In the first type, each speaker takes turn to speak during the conversation and the role changing happens when a speaker finishes his/her speech. 
In the second type, mixing voices happen during the interruption of other speakers. 
From these grammars, we synthesize a video training set containing multiple speakers by (1) randomly selecting different videos in the single subject training set; (2) concatenating these videos sequentially (i.e. cooperative mode); (3) mixing their voices in a short time window and vertically concatenating the video frames (i.e. competitive mode).
In all cases, $S^k_m$ is set to the one who occupies the audio segment or the all segments of the whole video, accordingly.
The Audio-Visual Transformer is optimized as: 
\vspace{-4mm}
\begin{equation}
\footnotesize
    \mathcal{L} = \alpha_{align} \mathcal{L}_{align} + \alpha_{visual} \mathcal{L}_{Cyc\_Sync} + \alpha_{audio} \mathcal{L}_{audio}
\end{equation}
where $\{\alpha_{align}, \alpha_{visual}, \alpha_{audio}\}$ are the parameters controlling their relative importance.


\section{Main Speaker Dataset}
While most of previous speaker datasets \cite{LRS2, LRS3, Columbia, roth2020ava} are mainly designed for the active speaker detection task, in this work, we further introduce a large-scale dataset for the Main Speaker Detection task. 
The proposed dataset 
is collected with three conversation types, i.e. \textit{discussion panel}, \textit{tele-conference}, and \textit{debate}, from several Youtube channels. 
Particularly, in the discussion panel videos, speakers take turn to speak during the conversation cooperatively. 
For the second type, videos consist of multiple subject (i.e. 3-5 people) talking through Skype or Zoom in a tele-conference. The third type is more challenging with debate-style videos where there are more interruptions among the two speakers in both cooperative and competitive manners.
For each collected video, we select segments of various lengths (i.e. from 6 seconds to 20 seconds) that can represent the property of the corresponding conversation style. 
%
In total, the dataset consists of 300 minutes videos. All clips are converted to have 25fps and 16kHz. The bounding box of the main speaker is also annotated. 

\vspace{-1mm}
\section{Experimental Results}

\begin{table}[t]
    \centering
    \footnotesize
    \caption{\textbf{Main Speaker Audio Separation on LRS2 with and without Domain Alignment}. The higher value is better.}
    \begin{tabular}{|c|c|c|c|}
    \hline
    \multicolumn{2}{|c|}{}         & \textbf{Ours w/o $\mathcal{L}_{align}$} &\textbf{ Ours W $\mathcal{L}_{align}$} \\
    \hline
    \multirow{3}{*}{\textbf{SDR (dB)}}  & \textbf{1S+N} &    14.4     &    \textbf{15.8}       \\
    \cline{2-4}
                      & \textbf{2S }  &    9.8       &      \textbf{10.3}     \\
    \cline{2-4}                    
                      & \textbf{2S+N} &    7.0      &     \textbf{7.2}      \\
    \hline
    \multirow{3}{*}{\textbf{PESQ}} & \textbf{1S+N} &     3.1        &    \textbf{3.3}       \\
    \cline{2-4}
                      & \textbf{2S }  &      2.7       &    \textbf{2.9}       \\
    \cline{2-4}
                      &\textbf{ 2S+N }&       2.5      &    \textbf{2.5}      \\
    \hline
    \end{tabular}
    \label{tab:AudioSeparation_Align}
    \vspace{-6mm}
\end{table}

\begin{table*} [!t]
	\footnotesize 
	\centering
	\caption{\textbf{Main Speaker Localization Accuracy with Cooperative Turn-Taking Conversation} (\%). For LRS2 and LRS3, a localization is considered correct if it lies within the true bounding box. For Columbia, F1 score is adopted.} 
	 \begin{tabular}{|l|cc|c|ccccc|}
    \hline
     \multirow{3}{*}{}&\multicolumn{2}{c|}{\textbf{Single Speaker}} & \multicolumn{6}{c|}{\textbf{Multiple Speakers}}\\
     \cline{2-9}
      \multirow{2}{*}{} & \multirow{2}{*}{\textbf{LRS2}} & \multirow{2}{*}{\textbf{LRS3}} & \multirow{1}{*}{\textbf{Columbia}} & \multicolumn{5}{c|}{\textbf{Columbia} (Per subject)}\\
      & \multicolumn{2}{c|}{}& \textbf{(Avg)} & \textbf{Bell} & \textbf{Boll} & \textbf{Lieb} & \textbf{Long} & \textbf{Sick}\\
    
    \hline
    Baseline (Random Pixel) & 2.8\% & 2.9\% & 8.5\% &7.8\%  &8.6\% &9.9\% &7.9\% &8.7\%\\
    Baseline (Center Pixel) & 23.9\% & 25.9\% & 14.9\% &13.0\%  &11.5\%  &21.4\% &19.2\% &17.9\%\\
    \hline
    Multisensory \cite{owens2018audio} & 99.3\% & 24.8\% & 52.7\% &  52.0\%& 43.8\% &62.3\% &64.8\% &60.9\%\\
    Chakravarty et al.\cite{chakravarty2016cross}  & $-$ & $-$& 80.2\% & 82.9\%& 65.8\%& 73.6\%& 86.9\%& 81.8\%\\
    
    SyncNet \cite{chung2016out}& $-$ & $-$ & 89.5\% &  93.7\%& 83.4\%& 86.8\%& \textbf{97.7\%}& 86.1\%\\
	LWTNet \cite{Afouras20b} & 99.6\%& 99.7\%& 90.8\%& 92.6\%& 82.4\%& 88.7\%& 94.4\%&95.9\%\\
	\hline
	\hline
    (A) \textbf{Ours - Block attention} & 99.7\%& 99.8\%& 92.7\% & 93.7\% & 85.0\% & 87.5\% & 92.8\% &  97.2\%\\
    (B) \quad +  across segments& \textbf{99.8\%}  & \textbf{99.9\%} & 93.4\% & 95.8\% &  85.0\% & 87.5\% & 92.8\% & 97.2\% \\
     \hline
    (C) \textbf{Ours - Speaker Attention \footnotemark} & 100\% & 100\% & 93.8\% &  95.8\% & 85.0\% & 91.6\% & 92.8\% & 97.2\%\\
	(D) \quad +  across segments & \textbf{100\%} & \textbf{100\%} & \textbf{94.9\%} &  \textbf{95.8\%} & \textbf{88.5\%} & \textbf{91.6\%} & 96.4\% & \textbf{97.2\%} \\
     \hline
\end{tabular}
	\label{tab:InturnLocalization}
	\vspace{-2mm}
\end{table*}

\noindent
\textbf{Data Setting.} Our training data include 29 hours of training videos from Lip Reading Sentences 2 (LRS2) \cite{LRS2}, and synthetic videos obtaining as presented in Sect. \ref{sec:LearningProcess}.
The length of synthetic segments varies from 4s to 8s decomposed into 2s short segments.
The overlapped ratio of the mixing voice is set to $\frac{1}{3}$.
For validation, we adopt the testing set of LRS2, Lip Reading Sentences 3 (LRS3) \cite{LRS3}, Columbia \cite{Columbia}, and our collected dataset.
While LRS2 and LRS3 include 0.5-hour to 1-hour testing videos, Columbia includes an 86-minute panel discussion. 
We adopt the ground truth for each active speaker in Columbia while annotating
bounding boxes using face detection \cite{deng2019retinaface} for LRS2 and LRS3.

\noindent
\textbf{Audio Data Preprocess.} We employ Short Time Fourier Transform (STFT) to represent the audio signal. Our STFT use Han window function which generates magnitude and phase of spectrograms. 
We set the hop length of 10 ms with a window length of 40ms at a sample rate of 16000Hz.

\noindent
\textbf{Visual Data Preprocess.} All training videos are re-sampled to a resolution of $160 \times 160$ pixels at $25$ FPS. This chosen resolution results in a feature map composing $N = 6 \times 6$ blocks.
During testing phase, we only re-sample the input video to $25$ FPS and retain the original resolution.

\noindent
\textbf{Network Architectures.} We employ 3D VGG-style network for visual deep feature embedding $F_v$, and 2D VGG-style network for audio embedding $F_a$.
The linear projections $\{\Omega_v^Q, \Omega_v^K, \Omega_v^V, \Omega_a^Q, \Omega_a^K, \Omega_a^V, \Omega^Q, \Omega^K, \Omega^V \}$ are implemented as the fully connected layers that project features to $512-D$ spaces. The mapping functions $\{\eta_v, \eta_a, \eta\}$ are implemented as residual-style MLP
consisting of 2 fully connected layers followed by the normalization layer \cite{ba2016layer} (the dimension of hidden layers is set to $1024$).
The audio-visual mask generator $D$ is implemented by a stack of 2 fully connected layers, which predicts both the magnitude mask and the phase mask of the spectrogram. We use the RetinaFace \cite{deng2019retinaface} 
for face detection widely used in face recognition 
\cite{Chen_FG2011, duong2019shrinkteanet, 9185981, Duong_ICASSP2011, Le_JPR2015, Luu_BTAS2009, Luu_FG2011, Luu_IJCB2011, 9108692}.

\begin{table*}[t]
	\footnotesize
	\begin{minipage}[t]{0.41\textwidth}
	\centering
	
	\footnotesize
    	\caption{\textbf{Main Speaker Localization Accuracy in Competitive Turn-Taking Conversation.}} 
    	 \begin{tabular}{|l|ccc|}
            \hline
            \multirow{1}{*}{} & \multirow{1}{*}{\textbf{Discussion}} & \multirow{1}{*}{\textbf{Tele}} & \multirow{2}{*}{\textbf{Debate}} \\
              \multirow{1}{*}{} & \multirow{1}{*}{\textbf{Panel}} &  \textbf{Conf}& \\
            
            \hline
            Baseline (Random) &4.8\%  &3.0\%  & 9.5\%\\
a            Baseline (Center) &1.3\%  &1.1\%  & 5.9\%\\
            \hline
           
             LWTNet\cite{Afouras20b}+large\_mag & 62.8\% & 55.07\%  & 58.7\% \\
             LWTNet\cite{Afouras20b}+high\_corr & 88.0\% & 80.0\%  &63.3\% 
            \\
        	\hline
            \textbf{Ours} & \textbf{90.2\%} & \textbf{83.3\%} & \textbf{69.4\%}\\
             \hline
        \end{tabular}
        \label{table:InterruptingConversation}
	\end{minipage} 
	\hspace{0.15cm}
	\begin{minipage}[t]{0.56\textwidth}
	\centering
   	\footnotesize
    	\caption{\textbf{Main Speaker Audio Separation on LRS2.}}
    	 \label{tab:AudioSeparation}
    	 \begin{tabular}{|l|ccc|ccc|} 
        \hline
         \multirow{3}{*}{}&\multicolumn{3}{c|}{\textbf{SDR (dB)$\boldsymbol\uparrow$}}&\multicolumn{3}{c|}{\textbf{PESQ}$\boldsymbol\uparrow$} \\ 
         \cline{2-7}
          \multirow{1}{*}{} & \multirow{1}{*}{\textbf{1S+N}} & \multirow{1}{*}{\textbf{2S}} & \multirow{1}{*}{\textbf{2S+N}} & \multirow{1}{*}{\textbf{1S+N}} & \multirow{1}{*}{\textbf{2S}} & \multirow{1}{*}{\textbf{2S+N}} \\
        \hline
        Mix input &1.3&1.31 &0.6 &1.1&1.1 &1.0  \\ 
        \hline
        SoundOfPixel \cite{Zhao_2018_ECCV} &9.4  & 1.5  &0.5 &1.2  &1.1  &1.0  \\ 
        Deep-Clustering \cite{hershey2016deep} & 9.0 & 6.0 & 3.2 & 2.3 & 2.3 & 1.9 \\ 
        Conv-TasNet \cite{PMID:31485462} & $-$ & 10.7 & $-$ & $-$ & $-$ & $-$ \\
        LWTNet \cite{Afouras20b} & $-$ &10.8  & $-$ & $-$ &3.0  & $-$  \\
    	\hline
    	\hline
    	\textbf{Ours (Audio Only)} & 11.1  & 9.1 & 7.0 & 2.8 & 2.8 & 2.5 \\ 
    	\quad +  across segments & 11.2  & 9.4 & 7.1 & 2.8 & 2.9 & 2.6  \\
        \hline
    	(A) \textbf{Ours - Block Attention} & 15.8 & 10.3 & 7.2 & 3.3 & 2.9 & 2.5  \\
    	(B) \quad +  across segments & 16.6 & 11.5 & 8.1 & \textbf{3.6} & 3.1 & 2.7  \\
    
    	(C) \textbf{Ours - Speaker Attention} & 16.5 & 10.5 & 7.5 & 3.4 & 3.0 &  3.0  \\
    	(D) \quad +  across segments & \textbf{16.7} & \textbf{11.6} & \textbf{8.2} & 3.4 & \textbf{3.1} & \textbf{3.1} \\
         \hline
        \end{tabular}
	\end{minipage}
	\vspace{-4mm}
\end{table*}
\noindent
\textbf{Model Configurations.} Our framework is implemented in PyTorch \cite{paszke2019pytorch} and all the models are trained on a machine with four NVIDIA P6000 GPUs. The batch size is set to 32 for each GPU. We use RMSProp optimizer with the started learning rate of 0.0001. 
We set the control parameters to $1.0$, i.e $\alpha_{align} = \alpha_{visual} = \alpha_{audio} = 1.0$. 

\begin{figure}[t]
    \centering
    \includegraphics[width=0.47\textwidth]{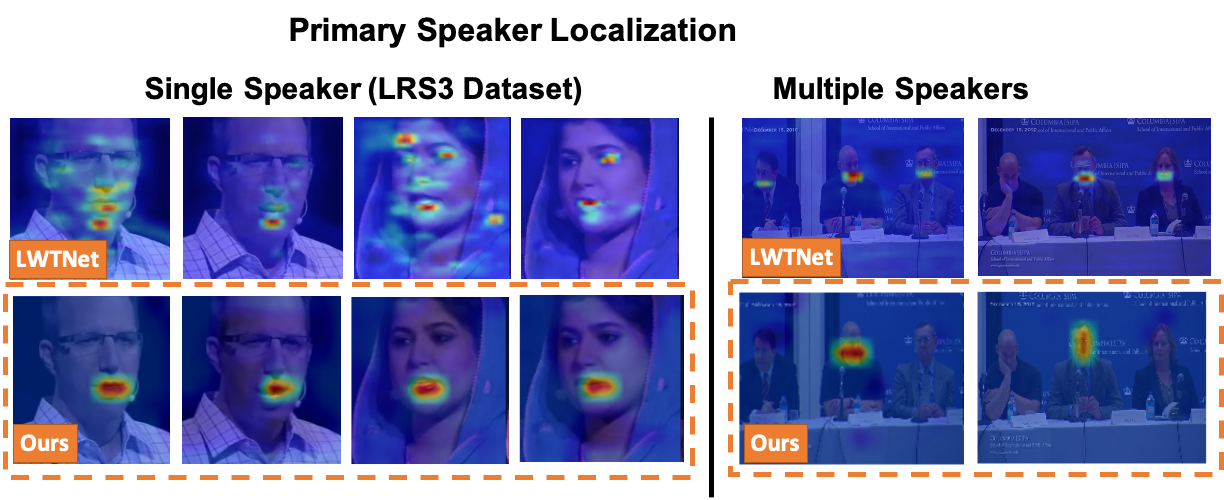}
    \caption{\textbf{Main Speaker Localization.}
    Visualization of attention mask localizing the main speaker. (\textbf{Best viewed in color})}
    \label{fig:visualize_primary_speaker_localization}
    \vspace{-6mm}
\end{figure}

\noindent
\textbf{Evaluation Metrics.} To compare against prior methods, we adopt four common metrics for localization and audio separation tasks. 
For single speaker videos, a localization is correct if its lies in the ground-truth bounding box of Main speakers. For multiple-speaker videos, F1 score is adopted for validation. 
To evaluate Main speaker separation, we adopt the protocol of multi-source speaker audio separation, and estimate the Signal-to-Distortion-Ratio (SDR) \cite{fevotte2005bss_eval} and Perceptual Evaluation of Speech Quality (PESQ) \cite{rix2001perceptual}.
\noindent
\textbf{Ablation Study.}
To study the effectiveness of our proposed cross-domain alignment method, we employ an ablation study with audio separation task 
on LRS2 using two configurations: without and with $\mathcal{L}_{align}$. 
We create synthetic testing video samples from LRS2 by combining audios from multiple videos. Three use-cases are evaluated including a primary  voice with background noise (1S+N); a primary voice mixed another speaker's voice (2S), and a primary voice mixed another speaker's voice plus background noise (2S+N). We report SDR (dB) and PESQ metrics for these cases in Table \ref{tab:AudioSeparation_Align}. By aligning the features of the two domains, the audio-visual correlations can be efficiently extracted and help to consistently improve SDR in all cases.








\vspace{-2mm}
\subsection{Main Speaker Localization}
\vspace{-2mm}



\noindent
\textbf{\textit{Cooperative Turn-Taking Conversation.}} In this type, as each speaker takes his/her turn to join the conversation, the main speaker is also the one who is actively speaking during the conversation. The localization accuracy for both single-speaker and multiple-speaker conversations in comparison to previous Active Speaker Detection approaches is reported in Table \ref{tab:InturnLocalization}. 
For each training mode of our model, we also include the configurations that take into account the correlations within and across segments. 
As can be seen, with the attention mechanisms as well as the domain alignment process, the visual and audio features are better correlated and provide more accurate locations of the main speaker. Moreover, when the spatial-temporal dimension is adopted in configuration (B) and (D), the performance is further boosted. Thanks to the correlations across segments (shown in Fig. \ref{fig:Visual_Visual_attention}), the location of each speaker is highly correlated with face of the same subject in other segments and, therefore, enable the tracking consistency of that speaker during the conversation. 
Our approach outperforms LWTNet \cite{Afouras20b} in all datasets with the margins from $0.1\%$ to $4.1\%$.
Fig. \ref{fig:visualize_primary_speaker_localization} shows our  
localization results compared to LWTNet.




\noindent
\textbf{\textit{Competitive Turn-Taking Conversation.}}
This type is more challenging as two speakers may speak at the same time. 
Therefore, although the two speakers can be both active speakers, only one of them is considered as the main speaker while the other one is the interrupter. 
For this task, beside the Random Pixel and Center Pixel baselines, we consider two additional localization strategies. We firstly employ LWTNet \cite{Afouras20b} to localize all active speakers of each video segment and then choose the main speaker as the one with (1) larger audio magnitude (i.e. \textit{large\_mag}), and (2) maximal audio-visual correlation  (i.e. \textit{high\_corr}).
Table \ref{table:InterruptingConversation} reports the localization accuracy on our collected dataset in terms of F1 score against the four baseline approaches.
These results again emphasizes the advantages of 
our proposed approach in the capability of automatically and robustly localize the main speaker in a conversation. 
The achieved improvements comes from three properties of the proposed model: (1) the present of the contextual attention from both visual and audio domains; (2) the domain feature alignment, and (3) the Cycle Synchronization Loss $\mathcal{L}_{Cyc\_Sync}$ that minimizes the disparity between the localization masks obtained from mixed voices and clean voice. 




\footnotetext{We report the accuracy of the face detection in single-speaker case.}

\vspace{-1mm}
\subsection{Main Speaker Audio Separation}
\vspace{-1mm}
To quantitatively evaluate the capability of audio separation for the proposed 
approach, we employ the evaluation protocol of \cite{Afouras20b} and use SDR and PESQ as the validation metrics. Similar to the previous section, we create synthetic testing videos from LRS2 on three cases, i.e. 1S + N, 2S, and 2S + N, and evaluate different configurations of our approach in comparison to previous methods as shown in Table \ref{tab:AudioSeparation}. 
%
%
With the spatial-temporal attentions, all configurations that take into account the cross-segment correlations get improvements from $0.3$ to $1.2$dB of SDR when separating voices of two speakers. Furthermore, the audio-visual attentions also give more cues to improve the separation process. 
We validate the roles of SROI by adopting two strategies (see Sect. \ref{sec:Visual_Self_Att}), i.e. block decomposition and face detection. Although the use of face detection can give more focus on face regions and produce further improvements, the block decomposition approach can still attend to the track of the same speaker across segments and give competitive performance. Moreover, our approach with both configurations outperforms LWTNet \cite{Afouras20b} in SDR and PESQ.

\vspace{-2mm}
\section{Conclusion}
\vspace{-2mm}
This work has presented a novel Audio-Visual Transformer approach for Main Speaker Localization and Audio Separation. Thanks to the introduced attention mechanisms in spatial-temporal dimension 
together with the domain alignment for better synchronization, our method can effectively localize and highlight the main speaker in both visual and audio channels on multi-speaker conversation videos.
Experiments in visual localization and audio separation tasks have shown the advantages of our proposal. 

\vspace{-5mm}
\paragraph{Acknowledgement.} This work is supported by NSF Data Science, Data Analytics that are Robust' and Trusted (DART) and Chancellor's Innovation Fund, UAF. It was also funded in part by the Arkansas Biosciences Institute, the agricultural and biomedical research program of the Arkansas Tobacco Settlement Proceeds Act of 2000.

{\small
\bibliographystyle{ieee_fullname}
\bibliography{egpaper_final}

\begin{thebibliography}{10}\itemsep=-1pt

\bibitem{LRS2}
Triantafyllos Afouras, Joon~Son Chung, Andrew Senior, Oriol Vinyals, and Andrew
  Zisserman.
\newblock Deep audio-visual speech recognition.
\newblock {\em TPAMI}, page 1–1, 2019.

\bibitem{afouras2018conversation}
T. Afouras, J.~S. Chung, and A. Zisserman.
\newblock The conversation: Deep audio-visual speech enhancement.
\newblock In {\em INTERSPEECH}, 2018.

\bibitem{LRS3}
T. Afouras, J.~S. Chung, and A. Zisserman.
\newblock Lrs3-ted: a large-scale dataset for visual speech recognition.
\newblock In {\em arXiv:1809.00496}, 2018.

\bibitem{afouras2019my}
Triantafyllos Afouras, Joon~Son Chung, and Andrew Zisserman.
\newblock My lips are concealed: Audio-visual speech enhancement through
  obstructions.
\newblock In {\em INTERSPEECH}, 2019.

\bibitem{Afouras20b}
Triantafyllos Afouras, Andrew Owens, Joon~Son Chung, and Andrew Zisserman.
\newblock Self-supervised learning of audio-visual objects from video.
\newblock In {\em ECCV}, 2020.

\bibitem{arandjelovic2017look}
Relja Arandjelovic and Andrew Zisserman.
\newblock Look, listen and learn.
\newblock In {\em ICCV}, 2017.

\bibitem{arandjelovic2018objects}
Relja Arandjelovic and Andrew Zisserman.
\newblock Objects that sound.
\newblock In {\em ECCV}, 2018.

\bibitem{ba2016layer}
Jimmy~Lei Ba, Jamie~Ryan Kiros, and Geoffrey~E Hinton.
\newblock Layer normalization.
\newblock {\em arXiv:1607.06450}, 2016.

\bibitem{barzelay2007harmony}
Zohar Barzelay and Yoav~Y Schechner.
\newblock Harmony in motion.
\newblock In {\em CVPR}, 2007.

\bibitem{Columbia}
Punarjay Chakravarty and Tinne Tuytelaars.
\newblock Cross-modal supervision for learning active speaker detection in
  video.
\newblock In {\em ECCV}, 2016.

\bibitem{chakravarty2016cross}
Punarjay Chakravarty and Tinne Tuytelaars.
\newblock Cross-modal supervision for learning active speaker detection in
  video.
\newblock In {\em ECCV}, 2016.

\bibitem{Chen_FG2011}
C. Chen, W. Yang, Y. Wang, K. Ricanek, and K. Luu.
\newblock Facial feature fusion and model selection for age estimation.
\newblock In {\em FG}, 2011.

\bibitem{chung2019said}
Joon~Son Chung, Bong{-}Jin Lee, and Icksang Han.
\newblock Who said that?: Audio-visual speaker diarisation of real-world
  meetings.
\newblock In Gernot Kubin and Zdravko Kacic, editors, {\em INTERSPEECH}, 2019.

\bibitem{chung2016out}
Joon~Son Chung and Andrew Zisserman.
\newblock Out of time: automated lip sync in the wild.
\newblock In {\em ACCV}, 2016.

\bibitem{cutler2000look}
Ross Cutler and Larry Davis.
\newblock Look who's talking: Speaker detection using video and audio
  correlation.
\newblock In {\em ICME}, 2000.

\bibitem{cutrone2019profiling}
Pino Cutrone.
\newblock Profiling performances of l2 listenership: Examining the effects of
  individual differences in the japanese efl context.
\newblock {\em TESOL}, 2019.

\bibitem{deng2019retinaface}
Jiankang Deng, Jia Guo, Yuxiang Zhou, Jinke Yu, Irene Kotsia, and Stefanos
  Zafeiriou.
\newblock Retinaface: Single-stage dense face localisation in the wild.
\newblock {\em arXiv:1905.00641}, 2019.

\bibitem{9054376}
Y. {Ding}, Y. {Xu}, S.~X. {Zhang}, Y. {Cong}, and L. {Wang}.
\newblock Self-supervised learning for audio-visual speaker diarization.
\newblock In {\em ICASSP}, 2020.

\bibitem{7780991}
Chi~Nhan Duong, Khoa Luu, Kha~Gia Quach, and Tien~D. Bui.
\newblock Longitudinal face modeling via temporal deep restricted boltzmann
  machines.
\newblock In {\em CVPR}, 2016.

\bibitem{10.1007/s11263-018-1113-3}
Chi~Nhan Duong, Khoa Luu, Kha~Gia Quach, and Tien~D. Bui.
\newblock Deep appearance models: A deep boltzmann machine approach for face
  modeling.
\newblock {\em IJCV}, 2019.

\bibitem{duong2019shrinkteanet}
Chi~Nhan Duong, Khoa Luu, Kha~Gia Quach, and Ngan Le.
\newblock Shrinkteanet: Million-scale lightweight face recognition via
  shrinking teacher-student networks.
\newblock {\em arXiv:1905.10620}, 2019.

\bibitem{8953561}
Chi~Nhan Duong, Khoa Luu, Kha~Gia Quach, Nghia Nguyen, Eric Patterson, Tien~D.
  Bui, and Ngan Le.
\newblock Automatic face aging in videos via deep reinforcement learning.
\newblock In {\em CVPR}, 2019.

\bibitem{9185981}
Chi~Nhan Duong, Kha~Gia Quach, Ibsa Jalata, Ngan Le, and Khoa Luu.
\newblock Mobiface: A lightweight deep learning face recognition on mobile
  devices.
\newblock In {\em BTAS}, 2019.

\bibitem{Duong_ICASSP2011}
Chi~Nhan Duong, Kha~Gia Quach, Khoa Luu, Hoai~Bac Le, and Karl~Ricanek Jr.
\newblock Fine tuning age estimation with global and local facial features.
\newblock In {\em ICASSP}, 2011.

\bibitem{10.1007/s11263-019-01165-5}
Chi~Nhan Duong, Kha~Gia Quach, Khoa Luu, T.~Hoang Le, Marios Savvides, and
  Tien~D. Bui.
\newblock Learning from longitudinal face demonstration--where tractable deep
  modeling meets inverse reinforcement learning.
\newblock {\em IJCV}, 2019.

\bibitem{8237665}
Chi~Nhan Duong, Kha~Gia Quach, Khoa Luu, T.~Hoang~Ngan Le, and Marios Savvides.
\newblock Temporal non-volume preserving approach to facial age-progression and
  age-invariant face recognition.
\newblock In {\em ICCV}, 2017.

\bibitem{Duong_2020_CVPR}
Chi~Nhan Duong, Thanh-Dat Truong, Khoa Luu, Kha~Gia Quach, Hung Bui, and
  Kaushik Roy.
\newblock Vec2face: Unveil human faces from their blackbox features in face
  recognition.
\newblock In {\em CVPR}, June 2020.

\bibitem{ephrat2018looking}
Ariel Ephrat, Inbar Mosseri, Oran Lang, Tali Dekel, Kevin Wilson, Avinatan
  Hassidim, William~T. Freeman, and Michael Rubinstein.
\newblock Looking to listen at the cocktail party: A speaker-independent
  audio-visual model for speech separation.
\newblock {\em ACM Trans. Graph.}, 2018.

\bibitem{fevotte2005bss_eval}
C{\'e}dric F{\'e}votte, R{\'e}mi Gribonval, and Emmanuel Vincent.
\newblock Bss\_eval toolbox user guide--revision 2.0.
\newblock 2005.

\bibitem{fisher2000learning}
John~W Fisher~III, Trevor Darrell, William Freeman, and Paul Viola.
\newblock Learning joint statistical models for audio-visual fusion and
  segregation.
\newblock {\em NIPS}, 2000.

\bibitem{gabbay2018seeing}
Aviv Gabbay, Ariel Ephrat, Tavi Halperin, and Shmuel Peleg.
\newblock Seeing through noise: Visually driven speaker separation and
  enhancement.
\newblock In {\em ICASSP}, pages 3051--3055, 2018.

\bibitem{gao2018learning}
Ruohan Gao, Rogerio Feris, and Kristen Grauman.
\newblock Learning to separate object sounds by watching unlabeled video.
\newblock In {\em ECCV}, 2018.

\bibitem{gao20192}
Ruohan Gao and Kristen Grauman.
\newblock 2.5 d visual sound.
\newblock In {\em CVPR}, 2019.

\bibitem{gao2019co}
Ruohan Gao and Kristen Grauman.
\newblock Co-separating sounds of visual objects.
\newblock In {\em ICCV}, 2019.

\bibitem{harwath2018jointly}
David Harwath, Adria Recasens, D{\'\i}dac Sur{\'\i}s, Galen Chuang, Antonio
  Torralba, and James Glass.
\newblock Jointly discovering visual objects and spoken words from raw sensory
  input.
\newblock In {\em ECCV}, 2018.

\bibitem{He_2017_ICCV}
Kaiming He, Georgia Gkioxari, Piotr Dollar, and Ross Girshick.
\newblock Mask r-cnn.
\newblock In {\em ICCV}, 2017.

\bibitem{hershey1999audio}
J Hershey and JR Movellan.
\newblock Audio-vision: Locating sounds via audio-visual synchrony.
\newblock {\em NIPS}, 1999.

\bibitem{hershey2016deep}
John~R Hershey, Zhuo Chen, Jonathan Le~Roux, and Shinji Watanabe.
\newblock Deep clustering: Discriminative embeddings for segmentation and
  separation.
\newblock In {\em ICASSP}, 2016.

\bibitem{hu2019deep}
Di Hu, Feiping Nie, and Xuelong Li.
\newblock Deep multimodal clustering for unsupervised audiovisual learning.
\newblock In {\em CVPR}, 2019.

\bibitem{NEURIPS2020_7288251b}
Di Hu, Rui Qian, Minyue Jiang, Xiao Tan, Shilei Wen, Errui Ding, Weiyao Lin,
  and Dejing Dou.
\newblock Discriminative sounding objects localization via self-supervised
  audiovisual matching.
\newblock In H. Larochelle, M. Ranzato, R. Hadsell, M.~F. Balcan, and H. Lin,
  editors, {\em NIPS}, pages 10077--10087, 2020.

\bibitem{izadinia2012multimodal}
Hamid Izadinia, Imran Saleemi, and Mubarak Shah.
\newblock Multimodal analysis for identification and segmentation of
  moving-sounding objects.
\newblock {\em TMM}, 2012.

\bibitem{jin2009supervised}
Zhaozhang Jin and DeLiang Wang.
\newblock A supervised learning approach to monaural segregation of reverberant
  speech.
\newblock {\em TASLP}, 2009.

\bibitem{khan2013speaker}
Faheem Khan and Ben Milner.
\newblock Speaker separation using visually-derived binary masks.
\newblock In {\em Auditory-Visual Speech Processing (AVSP) 2013}, 2013.

\bibitem{khosravan2019attention}
Naji Khosravan, Shervin Ardeshir, and Rohit Puri.
\newblock On attention modules for audio-visual synchronization.
\newblock In {\em CVPRW}, 2019.

\bibitem{kidron2005pixels}
Einat Kidron, Yoav~Y Schechner, and Michael Elad.
\newblock Pixels that sound.
\newblock In {\em CVPR}, 2005.

\bibitem{korbar2018co}
Bruno Korbar.
\newblock Co-training of audio and video representations from self-supervised
  temporal synchronization.
\newblock 2018.

\bibitem{Le_JPR2015}
H.~N. Le, K. Seshadri, K. Luu, and M. Savvides.
\newblock Facial aging and asymmetry decomposition based approaches to
  identiﬁcation of twins.
\newblock {\em Journal of Pattern Recognition}, 2015.

\bibitem{liu2013source}
Qingju Liu, Wenwu Wang, Philip~JB Jackson, Mark Barnard, Josef Kittler, and
  Jonathon Chambers.
\newblock Source separation of convolutive and noisy mixtures using
  audio-visual dictionary learning and probabilistic time-frequency masking.
\newblock {\em TSP}, 2013.

\bibitem{PMID:31485462}
Yi Luo and Nima Mesgarani.
\newblock Conv-tasnet: Surpassing ideal time-frequency magnitude masking for
  speech separation.
\newblock {\em TASLP}, 2019.

\bibitem{Luu_BTAS2009}
K. Luu, T.~D. Bui, K.~Ricanek Jr., and C.~Y. Suen.
\newblock Age estimation using active appearance models and support vector
  machine regression.
\newblock In {\em BTAS}, 2009.

\bibitem{Luu_FG2011}
K. Luu, T.~D. Bui, and C.~Y. Suen.
\newblock Kernel spectral regression of perceived age from hybrid facial
  features.
\newblock In {\em FG}, 2011.

\bibitem{Luu_IJCB2011}
K. Luu, K. Seshadri, M. Savvides, T.~D. Bui, and C.~Y. Suen.
\newblock Contourlet appearance model for facial age estimation.
\newblock In {\em IJCB}, 2011.

\bibitem{makino2007blind}
Shoji Makino, Te-Won Lee, and Hiroshi Sawada.
\newblock {\em Blind speech separation}.
\newblock Springer, 2007.

\bibitem{nachmani2020voice}
Eliya Nachmani, Yossi Adi, and Lior Wolf.
\newblock Voice separation with an unknown number of multiple speakers.
\newblock In {\em Proceedings of the 37th international conference on Machine
  learning}, 2020.

\bibitem{Nguyen_2021_CVPR}
Xuan-Bac Nguyen, Duc~Toan Bui, Chi~Nhan Duong, Tien~D. Bui, and Khoa Luu.
\newblock Clusformer: A transformer based clustering approach to unsupervised
  large-scale face and visual landmark recognition.
\newblock In {\em CVPR}, 2021.

\bibitem{owens2018audio}
Andrew Owens and Alexei~A Efros.
\newblock Audio-visual scene analysis with self-supervised multisensory
  features.
\newblock In {\em ECCV}, 2018.

\bibitem{owens2018learning}
Andrew Owens, Jiajun Wu, Josh~H McDermott, William~T Freeman, and Antonio
  Torralba.
\newblock Learning sight from sound: Ambient sound provides supervision for
  visual learning.
\newblock {\em IJCV}, 2018.

\bibitem{paszke2019pytorch}
Adam Paszke, Sam Gross, Francisco Massa, Adam Lerer, James Bradbury, Gregory
  Chanan, Trevor Killeen, Zeming Lin, Natalia Gimelshein, Luca Antiga, et~al.
\newblock Pytorch: An imperative style, high-performance deep learning library.
\newblock In {\em NIPS}, 2019.

\bibitem{dyglip_cvpr21}
Kha~Gia Quach, Pha Nguyen, Huu Le, Thanh-Dat Truong, Chi~Nhan Duong, Minh-Triet
  Tran, and Khoa Luu.
\newblock Dyglip: A dynamic graph model with link prediction for accurate
  multi-camera multiple object tracking.
\newblock In {\em CVPR}, 2021.

\bibitem{radfar2007single}
Mohammad~H Radfar and Richard~M Dansereau.
\newblock Single-channel speech separation using soft mask filtering.
\newblock {\em TASLP}, 2007.

\bibitem{ramaswamy2020see}
Janani Ramaswamy and Sukhendu Das.
\newblock See the sound, hear the pixels.
\newblock In {\em WACV}, 2020.

\bibitem{reddy2007soft}
Aarthi~M Reddy and Bhiksha Raj.
\newblock Soft mask methods for single-channel speaker separation.
\newblock {\em TASLP}, 2007.

\bibitem{rix2001perceptual}
Antony~W Rix, John~G Beerends, Michael~P Hollier, and Andries~P Hekstra.
\newblock Perceptual evaluation of speech quality (pesq)-a new method for
  speech quality assessment of telephone networks and codecs.
\newblock In {\em ICASSP}, 2001.

\bibitem{roth2020ava}
Joseph Roth, Sourish Chaudhuri, Ondrej Klejch, Radhika Marvin, Andrew
  Gallagher, Liat Kaver, Sharadh Ramaswamy, Arkadiusz Stopczynski, Cordelia
  Schmid, Zhonghua Xi, et~al.
\newblock Ava active speaker: An audio-visual dataset for active speaker
  detection.
\newblock In {\em ICASSP}, 2020.

\bibitem{senocak2018learning}
Arda Senocak, Tae-Hyun Oh, Junsik Kim, Ming-Hsuan Yang, and In So~Kweon.
\newblock Learning to localize sound source in visual scenes.
\newblock In {\em CVPR}, 2018.

\bibitem{tian2018audio}
Yapeng Tian, Jing Shi, Bochen Li, Zhiyao Duan, and Chenliang Xu.
\newblock Audio-visual event localization in unconstrained videos.
\newblock In {\em ECCV}, 2018.

\bibitem{vay_sgw_2019}
Vayer Titouan, R\'{e}mi Flamary, Nicolas Courty, Romain Tavenard, and Laetitia
  Chapel.
\newblock Sliced gromov-wasserstein.
\newblock In {\em NIPS}. 2019.

\bibitem{9108692}
Dat~T. Truong, Chi Nhan~Duong, Khoa Luu, Minh-Triet Tran, and Ngan Le.
\newblock Domain generalization via universal non-volume preserving approach.
\newblock In {\em CRV}, 2020.

\bibitem{fi13070179}
Thanh-Dat Truong, Chi~Nhan Duong, Minh-Triet Tran, Ngan Le, and Khoa Luu.
\newblock Fast flow reconstruction via robust invertible n × n convolution.
\newblock {\em Future Internet}, 2021.

\bibitem{wang2005video}
Wenwu Wang, Darren Cosker, Yulia Hicks, S Saneit, and Jonathon Chambers.
\newblock Video assisted speech source separation.
\newblock In {\em ICASSP}, 2005.

\bibitem{zhao2019sound}
Hang Zhao, Chuang Gan, Wei-Chiu Ma, and Antonio Torralba.
\newblock The sound of motions.
\newblock In {\em ICCV}, 2019.

\bibitem{Zhao_2018_ECCV}
Hang Zhao, Chuang Gan, Andrew Rouditchenko, Carl Vondrick, Josh McDermott, and
  Antonio Torralba.
\newblock The sound of pixels.
\newblock In {\em ECCV}, 2018.

\bibitem{10.1007/978-3-030-58610-2_4}
Hang Zhou, Xudong Xu, Dahua Lin, Xiaogang Wang, and Ziwei Liu.
\newblock Sep-stereo: Visually guided stereophonic audio generation by
  associating source separation.
\newblock In {\em ECCV}, 2020.

\end{thebibliography}
}


\end{document}